\documentclass[conference]{IEEEtran}
\usepackage{graphicx}
\ifCLASSINFOpdf
\else
\fi
\usepackage{cite}
\usepackage{amsmath}

\begin{document}
%
\title{Deriving a Representative Vector for Ontology Classes with Instance Word Vector Embeddings}

\author{\IEEEauthorblockN{Vindula Jayawardana, Dimuthu Lakmal, Nisansa de Silva, Amal Shehan Perera,\\Keet Sugathadasa, Buddhi Ayesha}
\IEEEauthorblockA{Department of Computer Science \& Engineering\\University of Moratuwa}
\IEEEauthorblockA{Email: vindula.13@cse.mrt.ac.lk}
}



%


\maketitle

\begin{abstract}
Selecting a representative vector for a set of vectors is a very common requirement in many algorithmic tasks. Traditionally, the mean or median vector is selected. Ontology classes are sets of homogeneous instance objects that can be converted to a vector space by word vector embeddings. This study proposes a methodology to derive a representative vector for ontology classes whose instances were converted to the vector space. We start by deriving five candidate vectors which are then used to train a machine learning model that would calculate a representative vector for the class. We show that our methodology out-performs the traditional mean and median vector representations.\\
\end{abstract}


%
\IEEEpeerreviewmaketitle

\noindent\textbf{\textit{keywords}}: Ontology, Word Embedding, Representative Vector, Neural Networks, word2vec

\section{Introduction}
Semantic models are used to present hierarchy and semantic meaning of concepts. Among them ontologies are a widely used superlative model extensively applied to many fields. As defined by Thomas R. Gruber~\cite{Gruber1993}, an ontology is a ``formal and explicit specification of a shared conceptualization". The use of ontologies is becoming increasingly involved in various computational tasks given the fact that ontologies can overcome limitations in traditional natural language processing methods in domains such as text classification~\cite{textClassificationwithOntology,de2015safs3}, word set expansions~\cite{de2013semi}, linguistic information management~\cite{miller1990introduction,miller1990nouns,fellbaum1998wordnet, wijesiri2014building}, medical information management~\cite{huang2016omnisearch, huang2016development}, and Information Extraction~\cite{wimalasuriya2010ontology,de2017Discovering}.

However very few attempts have been made on representing ontology classes in different representations such as vector representations. The importance of having different representations for ontology classes is emphasized when it comes to ontology mapping, ontology merging, ontology
integration, ontology alignment and semi automated ontology population~\cite{ontoMap}. However sophisticated researches on representing ontology classes in different representations is still an open ended question.

In this study we propose a novel way of deriving representative vectors for ontology classes. This is an important problem in the domain of automatic ontology population and automatic ontology class labeling. We use a distributed representation of words in a vector space grouped together~\cite{mikolov2013distributed}, achieved by means of word vector embeddings to transform the word strings in instances and the class labels to the same vector space. For this task of word embedding, we chose the neural network based method: word2vec,  proposed by Tomas Mikolov et al.~\cite{mikolov2013efficient}, which is a model architecture for computing continuous vector representations of words from very large data sets. 


In the proposed methodology we created an ontology in the domain of consumer protection law with the help of legal experts. The word2vec model was trained with the legal cases from FindLaw~\cite{caseLaw} online database. The word embedding vectors of the instances and the class labels in the created ontology were the obtained using the trained word2vec model. For each ontology class a number of candidate vectors were then calculated using the word embedding vectors of the instances. The candidate vectors and the class label vectors were then used to train a machine learning model to predict the best representative vector. We show that our proposed methodology outperforms the traditional average (mean) vector representation and median vector representation in all classes. On average, the distance of our representative vector to the class vector is $0.82$, while the mean vector has a distance of $1.31$ and the median vector has a distance of $1.51$. Respectively, it is a $37\%$ and $50\%$ improvement.  



The rest of this paper is organized as follows: In Section~\ref{Background} we review previous studies related to this work. The details of our methodology for deriving a representative vector for ontology classes with instance word vector embeddings is introduced in Section~\ref{Methodology}. In Section~\ref{Results}, we demonstrate
that our proposed methodology produces superior 
results outperforming traditional
approaches. At last, we conclude and discuss some future works in Section~\ref{conclusion}.

\section{Background and Related Work}
\label{Background}
This section discusses the background details of the techniques utilized in this study and related previous studies carried out by others in various areas relevant to this research. The following subsections given below cover important key areas of this study.

\subsection{Ontologies}
In many areas, ontologies are used to organize information as a form of knowledge representation. An ontology may model either the world or a part of it as seen by the said area's viewpoint~\cite{de2013semi}. 

\textit{Individuals} (instances) make up the ground level of an ontology. These can be either concrete objects or abstract objects. Individuals are then grouped into structures called \textit{classes}. Depending on the domain on which the ontology is based, a class in an ontology can be referred to as a concept, type, category, or a kind. More often, the definition of a class and the role thereof is analogous to that of a collection of individuals with some additional properties that distinguish it from a mere set of objects. A class can either subsume, or be subsumed by, another class. This subsuming process give rise to the class hierarchy and the concept of super-classes (and sub-classes).    

\subsection{Word set expansion}
In many Natural Language Processing (NLP) tasks, creating and maintaining word-lists is an integral part. The said word-lists usually contain words that are deemed to be homogeneous in the level of abstraction involved in the application. Thus, two words $W_1$ and $W_2$ might belong to a single word-list in one application but belong to different word-lists in another application. This fuzzy definition and usage is what makes creation and maintenance of these word-lists a complex task.    

For the purpose of this study, we selected the algorithm presented in~\cite{de2013semi} which was built on the earlier algorithm described in~\cite{de2011semap}. The reason for this selection is: WordNet~\cite{miller1990introduction} based linguistic processes are reliable due to the fact that the WordNet lexicon was built on the knowledge of expert linguists.

\subsection{Word Embedding}
Word embedding systems, are a set of natural language modeling and feature learning techniques, where words from a domain are mapped to vectors to create a model that has a distributed representation of words, first proposed by~\cite{mikolov2013distributed}. Each of the words in a text document is mapped to a vector space. In addition to that, word meanings and relationships between the words are also mapped to the same vector space. word2vec\footnote{https://code.google.com/p/word2vec/}\cite{mikolov2013efficient}, GloVe~\cite{pennington2014glove}, and Latent Dirichlet Allocation (LDA)~\cite{das2015gaussian} are leading Word Vector Embedding systems. Both Word2vec and GloVe use \textit{word to neighboring word} mapping to learn dense embeddings. The difference is that Word2vec uses a neural network based approach while GloVe uses matrix factorization mechanism. LDA also has an approach that utilizes matrices but there the words are mapped with the relevant sets of documents. Due to the flexibility and ease of customization, we picked word2vec as the word embedding method for this study.

Word2vec is used in sentiment analysis~\cite{TwitterSentimental,sinaWeibo,chinessword2vec,sentimentalAnalysisInword2vec} and text classification~\cite{lilleberg2015support}. In the ontology domain there are two main works that involve word2vec: Gerhard Wohlgenannt et al.~\cite{word2VecOntologyPaper}'s approach to emulate a simple ontology using word2vec and Harmen Prins~\cite{word2VecOntologyPaper2}'s usage of word2vec extension: node2vec~\cite{node2VecOntologyPaper}, to overcome the problems in vectorization of an ontology.   





\subsection{Clustering}
Clustering is a seminal part in exploratory data mining and statistical data analysis. The objective of clustering is to group a set of items into separate sub-sets (clusters) where the items in a given cluster is more similar to each other than any of them is similar to an item from a different cluster. The used similarity measure and the desired number of clusters are application dependent. A clustering algorithm is inherently modeled as an iterative multi-objective optimization problem that involves trial and error which tries to move towards a state that exhibits the desired properties. Out of all the clustering methods available, we selected k-means clustering due to the easiness of implementation and configuration.  

Arguably, k-means clustering was first proposed by Stuart Lloyd~\cite{lloyd1982least} as a method of vector quantization for pulse-code modulation in the domain of signal processing. The objective is to partition $n$ observations into $k$ clusters where each observation $i$ belongs to the cluster with the mean $m$ such that in the set of cluster means $M$, $m$ is the closest to $i$ when measured by a given vector distance measure. It is implicitly assumed that the cluster mean serves as the prototype of the cluster. This results in the vector space being partitioned into Voronoi cells.

\subsection{Support Vector Machines}
Support Vector Machines~\cite{cortes1995support} is a supervised learning model that is commonly used in machine learning tasks that analyze data for the propose of classification or regression analysis. The SVM algorithm works on a set of training examples where each example is tagged as to be belonging to one of two classes. The objective is to find a hyperplane dividing these two classes such that, the examples of the two classes are divided by a clear gap which is as wide as mathematically possible. Thus the process is a non-probabilistic binary linear classifier task. The aforementioned gap is margined by the instances that are named \textit{support vectors}.


The idea of using SVMs for task pertaining ontologies are rather rare. However, a study by Jie Liu et al.~\cite{liu2012svm} defined a similarity cube which consists of similarity vectors combining similar concepts, instances and structures of two ontologies and then processed it through SVM based mapping discovery function to map similarities between two ontologies. Further, another study by Jie Liu et al.~\cite{liu2013ontology} has proposed a method of similarity aggregation using SVM, to classify weighted similarity vectors which are calculated using concept name and properties of individuals of ontologies. Our usage of SVM in the ontology domain in this paper is different from their approach and hence entirely novel.

In Generic SVM, new examples that are introduced are then predicted to be falling into either class depending on the relationship between that new example and the hyperplane that is dividing the two classes. However, in this study we do not need to employ the new instance assignment. We are only interested in calculating the support vectors. The usage and rationalization of this is given in Section~\ref{Method:svm}




\section{Methodology}
\label{Methodology}

\begin{figure*}[!htb]
\centering
\includegraphics[width=0.85\textwidth]{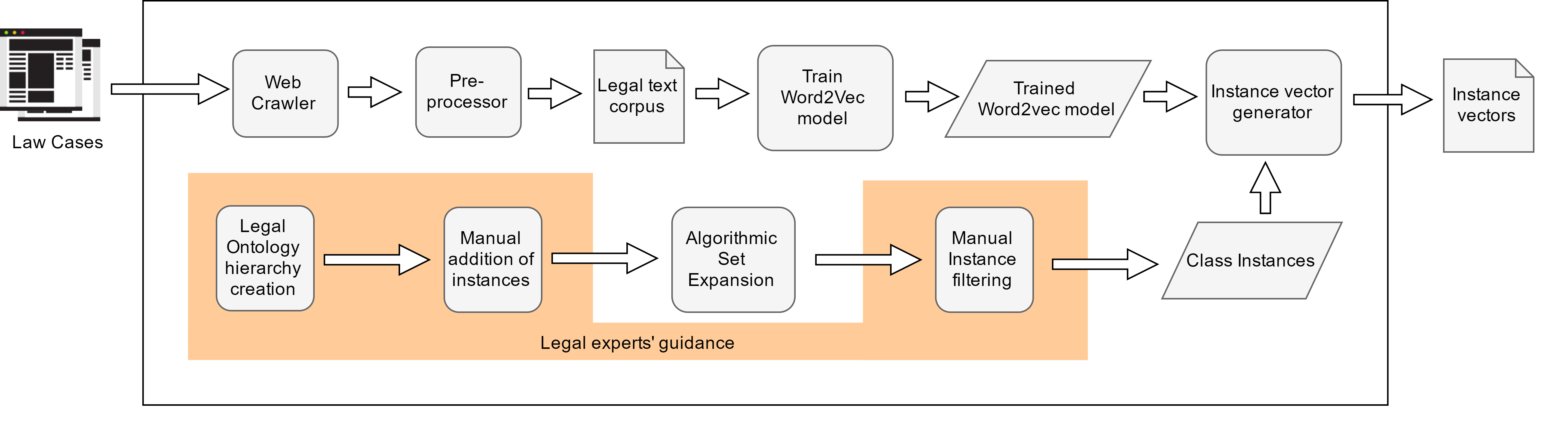}
\caption{Flow diagram of the methodology for deriving instance vectors}
\label{fig:processFlowAB}
\end{figure*}

\begin{figure*}[!htb]
\centering
\includegraphics[width=0.8\textwidth]{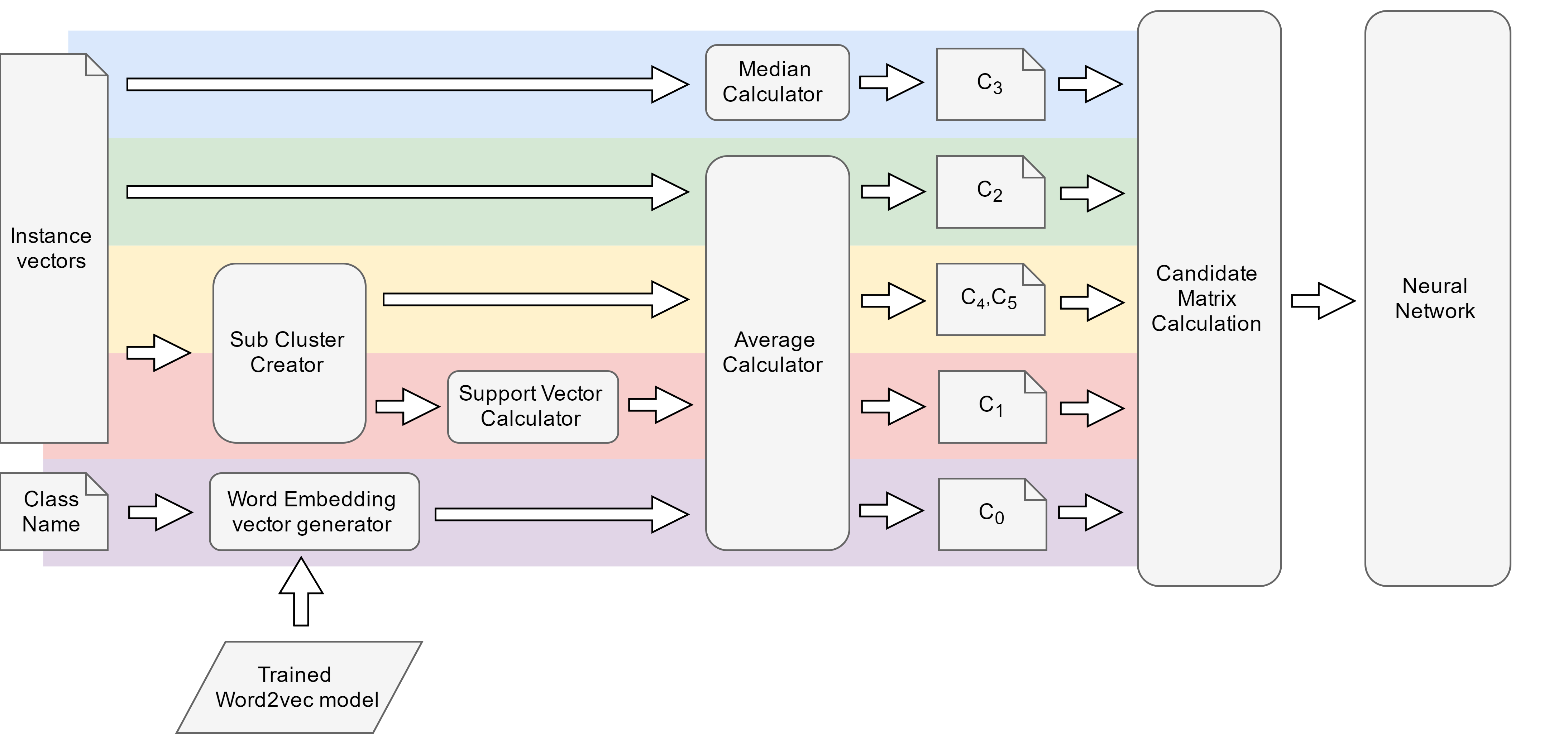}
\caption{Flow diagram of the methodology for training and testing the neural network}
\label{fig:processFlowCG}
\end{figure*}

We discuss the methodology that we used for deriving a vector representation for ontology classes using instance vector embeddings in this section. Each of the following subsections describe a step of our process. An overview of the methodology we propose in Sections~\ref{Method:onto} and~\ref{Method:wordEm} is illustrated in~\figurename~\ref{fig:processFlowAB} and an overview of the methodology we propose from Section~\ref{Method:SubCluster} to Section~\ref{ml} is illustrated in~\figurename~\ref{fig:processFlowCG}.

\subsection{Ontology Creation}
\label{Method:onto}

We created a legal ontology based on the consumer protection law, taking Findlaw~\cite{caseLaw} as the reference. After creating the ontology class hierarchy, we manually added seed instances for all the classes in the ontology. This was done based on manual inspection of the content of legal cases under consumer protection law in Findlaw. Next we used the algorithm proposed in~\cite{de2013semi} to expand the instance sets. The expanded lists were then pruned manually to prevent conceptual drift. This entire process was done under the supervision and guidance of experts from the legal domain. 

\subsection{Training word Embeddings}
\label{Method:wordEm}
The training of the word embeddings was the process of building a \textit{word2vec} model using a large legal text corpus obtained from Findlaw~\cite{caseLaw}. The text corpus consisted of legal cases under 78 law categories. In creating the legal text corpus we used \textit{Stanford CoreNLP} for preprocessing the text with tokenizing, sentence splitting, Part of Speech (PoS) tagging, and lemmatizing. 

The motive behind using a pipeline that pre-processes text up to and including lemmatization instead of the traditional approach of training the word2vec model with just tokenized text\cite{mikolov2013efficient}, was to map all inflected forms of a given lemma to a single entity. In the traditional approach each inflected form of a lemma gets trained as a separate vector. This dilutes the values that is extracted from the context of that lemma between all inflected forms. Thus sometimes resulting in values not meeting the threshold when a test of significance is done. By having all inflicted forms to be reduced to the relevant lemma and train, we made sure that all the contributions of the context of a given lemma is collected at a single vector, thus making the system more accurate. Secondly, having a unique vector for each inflected form makes the model unnecessarily large and heavy. This results in difficulties at both training and testing. Our approach of lemmatizing the corpus first solves that problem as well. In addition to this, to reduce ambiguities caused by the case of the strings, we converted all strings to lowercase before training the word2vec model. This too is a divergence from the conventional approach.   


\subsection{Sub-Cluster Creation}
\label{Method:SubCluster}

By definition, the instances in a given class of an ontology is more semantically similar to each other than instances in other classes. But no matter how coherent a set of items is, as long as that set contains more than one element, it is possible to create non-empty sub-sets that are proper subsets of the original set. This is the main motivation behind this step of our methodology. Following this rationale, it was decided that it is possible to find at least one main schism in the otherwise semantically coherent set of more than one instances. 

Given that even then, the schisms would be fairly small by definition in comparison to the difference of instances in one class and the instances of another class, it was decided to stop the sub-set creation at 2. Which means we decided to divide the instances in a single class in to two sub-clusters. For this purpose, we use K-means clustering with K=2. For an example, we are subdividing the "Judge" class using k-means to "Judge1" and "Judge2" and then use the support vectors between those two to predict the label of "Judge". The centers of these sub-clusters are two of the candidate vectors used in Section~\ref{Method:Candidate}.

\subsection{Support Vector Calculation}
\label{Method:svm}

It is clear that the specification of the problem handled in this study is more close to a clustering task than a classification task. In addition to that, given the fact that each time we would be running the proposed algorithm, it would be for instances in a single class. That implies that, even if we are to assign class labels to the instances, to model it as a classification task, it would have been a futile effort because there exist only one class. Thus, having a classifying algorithm such as Support Vector Machines (SVM) as part of the process of this study might seem peculiar at the beginning. However, this problem is no longer an issue due to the steps that were taken in the Section~\ref{Method:SubCluster}. Instead of the single unified cluster with the homogeneous class label, the previous step yielded two sub-clusters with two unique labels. Given that the whole premise of creating sub-clusters was based on the argument that there exists a schism in the individual vectors in the class, it is logical to have the next step to quantify that schism. 

For this task we used an SVM. The SVM was given the individuals in the two sub-clusters as the two classes and was directed to discover the support vectors. This process found the support vectors to be on either side of the schism that was discussed in Section~\ref{Method:SubCluster}. 

In identifying the support vectors, we used the algorithm used by Edgar Osuna et al.~\cite{SVM} in training support vector machines and then performed certain modifications to output the support vectors. 



\subsection{Candidate Matrix Calculation}
\label{Method:Candidate}

With the above steps completed, in order to derive the vector representation for ontology classes, we calculated a number of candidate vectors for each class and then derived the candidate matrix from those candidate vectors. We describe the said candidate vectors below:

\begin{itemize}
\item Average support vector ($C_{1}$)
\item Average instance vector ($C_{2}$) 
\item Class Median vector ($C_{3}$)
\item Sub-cluster average vectors ($C_{4}, C_{5}$)
\end{itemize}

The \textbf{Class name Vector ($C_0$)} is obtained by performing word vectorization on the selected class's class name and it was used as our desired output.

\subsubsection{Average support vector ($C_{1}$)}
We identified the support vectors which mark the space around the hyperplane that divides the class into the two subclasses as mentioned in Section~\ref{Method:svm}. We take the average of the said support vectors as the first candidate vector. The rationale behind this idea is that as described in Section~\ref{Method:SubCluster}, there exists a schism in between the two sub-classes. The support vectors mark the edges of that schism which means the average of the support vectors fall on the center of the said schism. Given that the schism is a representative property of the class, it is rational to consider this average support vector that falls in the middle of it as representative of the class. We averaged the instance vectors as shown in equation~\ref{Eq:VectorAverageFormula} to calculate the average support vector. 

\begin{equation}
C = \frac{\displaystyle \sum_{i=1}^{N} a_i V_i }{\displaystyle \sum_{i=1}^{N} a_i }
\label{Eq:VectorAverageFormula}
\end{equation}

Here, $N$ is the total number of vectors in the class. $V_i$ represents instance vectors. $a_i$ is the support vector membership vector such that: if the $i$th vector is a support vector, $a_i$ is $1$ and otherwise, it is $0$. Here $C$ is $C_{1}$ which represents the average support vector candidate vector.

\subsubsection{Average instance vector ($C_{2}$)}
This is by far the most commonly used class representation vector in other studies. We took all the instance vectors of the relevant class, and averaged them to come up with this candidate vector for the class. The Average instance vector was also calculated using the equation~\ref{Eq:VectorAverageFormula}. However, this time all the $a_i$s were initiated to $1$. In that case, $C$ is $C_{2}$ which represents the average instance vector.

\subsubsection{Class Median vector ($C_{3}$)}
Out of the instance vectors of a class, we calculated the median vector of them and added it as a candidate vector for that class. The Class Median vector ($C_{3}$) was calculated as shown in equation~\ref{Eq:VectorMedianFormula} where: $V$ is the set of instance vectors in the class. $v_i$ is the $i$th instance vector. $M$ is the number of dimensions in an instance vector. $x_{avg,j}$ is the $j$th element in the average instance vector $C_{2}$ that was calculated above using equation~\ref{Eq:VectorAverageFormula}.   

\begin{equation}
C_{3} =arg~\underset{v_i \in V}{\mathrm{min}} \Bigg\{ \sum_{j=1}^{M} {( \, x_{i,j}- x_{avg,j}) \,}^{2} \Bigg\}
\label{Eq:VectorMedianFormula}
\end{equation}

\subsubsection{Sub-cluster average vectors ($C_{4}, C_{5}$)}

We took all the instance vectors of one sub-cluster and averaged them to calculate $C_{4}$ and then did the same for the other sub-cluster to calculate $C_{5}$. The rationale behind this step is the fact that as described in Section~\ref{Method:SubCluster}, the two sub-clusters that we find are representative of the main division that exists in the class. Thus it is justifiable to consider the centroid of each of those sub-clusters. 

Each sub-cluster average instance vector was also calculated using the equation~\ref{Eq:VectorAverageFormula}. However, this time all the $a_i$s in the first cluster was initiated to $1$ and the $a_i$s in the second cluster was initiated to $0$. In that case, $C$ is $C_{4}$ which represents the average instance vector for the first cluster. Next the same function was used with all the $a_i$s in the first cluster initiated to $0$ and the $a_i$s in the second cluster initiated to $1$ to calculate $C$ which was assigned to $C_{5}$.
  
\subsubsection{Candidate Matrix}
Finally the candidate matrix $M_{cand}$ for each class is made by concatenating the transposes of $C_1$ though $C_5$ as shown in equation~\ref{Eq:candidateMatrix}.

\begin{equation}
M_{cand} = \Big\{C_{1}^{T},C_{2}^{T},C_{3}^{T},C_{4}^{T},C_{5}^{T} \Big\}
\label{Eq:candidateMatrix}
\end{equation}

\subsection{Optimization Goal}
\label{optimization}
After calculating the candidate vectors, we proposed an optimal vector that represents the given class based on the optimization goal as follows:

\begin{equation}
Y = \frac{\displaystyle \sum_{i=1}^{M} C_iW_i }{\displaystyle \sum_{i=1}^{M} W_i }
\label{Eq:optimizationFormular}
\end{equation}

Here, $Y$ is the predicted class vector for the given class. $M$ is the number of candidate vectors for a given class. $C_i$ and  $W_i$ represents the $i$th candidate vector and the associated weight of that candidate vector respectively. Here the $W_i$ is calculated using the method described in Section~\ref{ml}.

\subsection{Machine Learning implementation for weight calculation}
\label{ml}

The main motive behind adding a weight for each candidate vector is to account for the significance of the candidate vector towards the optimized class vector. We decided to use machine learning to calculate the weight vector. The machine learning method we used is a neural network.

The dataset is structured in the traditional $(X,D)$ structure where $X$ is the set of inputs and $D$ is the set of outputs. An input tuple $x$ (such that $x \in X$), has elements $x_{i,j}$. The matching output tuple $d$ (such that $d \in D$) has a single element. 

$X$ and $D$ is populated as follows: Take the $i$th row of the candidate matrix $M_{cand}$ of the class as $x$ and add it to $X$. Take the $i$th element of $C_{0}$ of the class and add it to $D$. Once this is done for all the classes, we get the $(X,D)$ set. It should be noted that since the weights are learned universally and not on a class by class basis, there will be one large dataset and not a number of small datasets made up of one per class. The reason for this is to make sure that the model does not overfit to one class and would instead generalize over all the data across the entire ontology. Because of this approach, we end up with a considerable amount of training data which again justifies our decision to use machine learning. For a word embedding vector length of $N$ over $M$ number of classes, this approach creates $NM$ number of training examples.

\section{Results}
\label{Results}
\begin{figure}[!htb]
\centering
\includegraphics[width=0.4\textwidth]{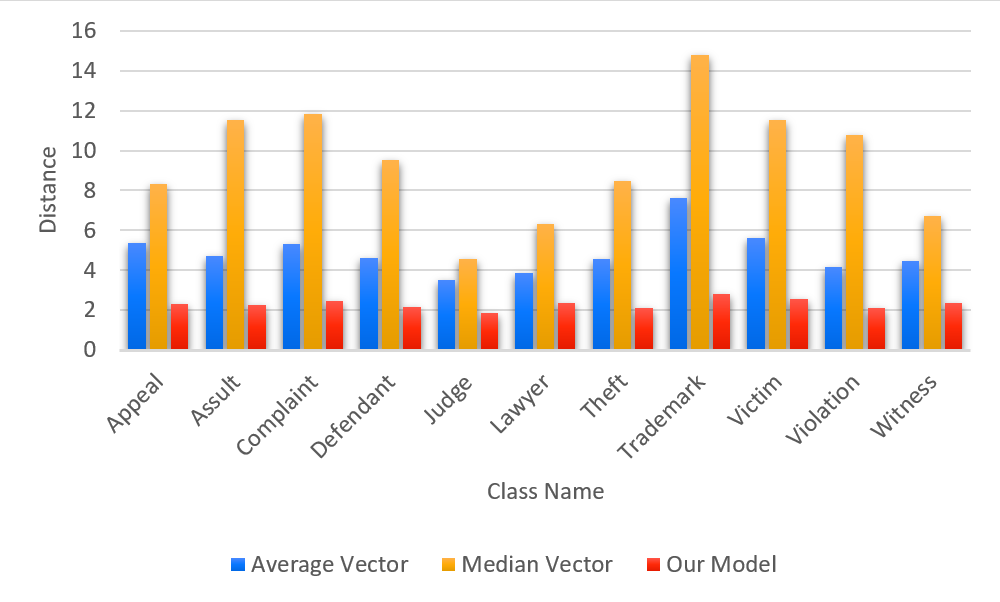}
\caption{Comparison of the distance measures of representative vectors}
\label{fig:comp}
\end{figure}

We used a set of legal ontology classes seeded by the legal experts and then expanded by the set expansion~\cite{de2013semi} algorithm under the guidance of the same legal experts. We report our findings below in the table~\ref{tab:res} and in\figurename\ref{fig:comp} we show a visual comparison of the same data. We illustrate the results that we obtained pertaining to ten prominent legal concept classes as well as the mean result across all the classes considered. We compare the representative vector proposed by us against the traditional representative vectors: average vector and median vector. All the results shown in the table are Euclidean distances obtained between the representative vector in question against the respective $C_{0}$ vector.   

\begin{table}[!htb]
	\centering
	\caption{Distance measures of the classes and the average over 10 classes }\label{tab:res}
	\begin{tabular}{|l|c|c|c|}
    	\hline
        & Average Vector & Median Vector & Our Model\\
		\hline
		Appeal & 5.37 & 8.32 & \textbf{2.31} \\
        \hline
        Assault & 4.73 & 11.53 & \textbf{2.27}\\
        \hline
        Complaint & 5.33 & 11.82 & \textbf{2.46}\\
        \hline
        Defendant & 4.60 & 9.51  & \textbf{2.15}\\
        \hline
        Judge & 3.52 & 4.56 & \textbf{1.84}\\
        \hline
        Lawyer & 3.89 & 6.33 & \textbf{2.35}\\
        \hline
        Theft & 4.56 & 8.49 & \textbf{2.13}\\
        \hline
        Trademark & 7.63 & 14.79 & \textbf{2.81}\\
        \hline
        Victim & 5.64 & 11.52 & \textbf{2.54}\\
        \hline
        Violation & 4.18 & 10.80 & \textbf{2.09}\\
        \hline
        Witness & 4.46 & 6.73 & \textbf{2.34}\\
        \hline
        \hline
        Class mean & 1.31 & 1.51 & \textbf{0.82}\\
        \hline
	\end{tabular}	
\end{table}

It can be observed from the results that the traditional approach of taking the average vector as the representative vector of a class is outperforming the other traditional approach of using the median vector. However, our proposed method outperforms both the average and median vectors in all cases. For an example, considering the "Judge" class, it can be seen that our model vector perform 47.8\% better than the average vector where it is 53.8\% better in the "Complaint" class.

\section{Conclusion and Future Works}
\label{conclusion}
In this work, we have demonstrated that the proposed method works as a better representation for a set of instances that occupy an ontology class than the traditional methods of using the average vector or the median vector. This discovery will be helpful in mainly two important tasks in the ontology domain. 

The first one is further populating an already seeded ontology. We, in this study used the algorithm proposed in~\cite{de2013semi} for this task to obtain a test dataset. However, that approach has the weakness of being dependent on the WordNet lexicon. A methodology built on the representative vector discovery algorithm proposed in this work will not have that weakness. This is because all the necessary vector representations are obtained from word vector embeddings done using a corpus relevant to the given domain. Thus all the unique jargon would be adequately covered without much of a threat of conceptual drift. As future work, we expect to take this idea forward for automatic ontology population.  

The second important task in the ontology domain that this method will be important is, class labeling. In this study we have demonstrated that our method is capable in deriving the closest vector representation to the class label. Thus, the converse of that would be true as well. That would be a \textit{topic modeling}~\cite{das2015gaussian} task. The idea is that if given an unlabeled class, the method proposed by this study can be used to derive the representative vector. Then by querying the word embedding vector space it is possible to obtain the most suitable class label (topic) candidate.  

\bibliographystyle{IEEEtran}
\bibliography{VectorRep}

\end{document}